\def\year{2021}\relax
\documentclass[letterpaper]{article} 
\usepackage{aaai21}  
\usepackage{times}  
\usepackage{helvet} 
\usepackage{courier}  
\usepackage[hyphens]{url}  
\usepackage{graphicx} 
\usepackage{natbib}  
\usepackage{caption} 
\usepackage{amsmath}
\usepackage{amsfonts}
\usepackage{amsthm}
\usepackage{bm}
\usepackage{algorithm}
\usepackage{algorithmic}
\usepackage{xspace}
\usepackage{enumitem}
\usepackage{multirow}
\usepackage{rotating}
\usepackage{subcaption}

\theoremstyle{definition}

\usepackage{xcolor}

\DeclareMathOperator*{\cat}{CAT}
\usepackage{graphicx}
\graphicspath{ {./images/} }

\title{Towards Sample Efficient Agents through Algorithmic Alignment (Student Abstract)}
\author{
    Mingxuan Li\thanks{Blog: https://mingxuan.me}, 
    Michael L. Littman,\\
}
\affiliations{
    Department of Computer Science, Brown University, 
    Providence, RI 02912-1910\\
    mingxuan\_li@brown.edu, mlittman@cs.brown.edu

}

\begin{document}

\maketitle

\begin{abstract}
In this work, we propose and explore Deep Graph Value Network (DeepGV) as a promising method to work around sample complexity in deep reinforcement-learning agents using a message-passing mechanism. The main idea is that the agent should be guided by structured non-neural-network algorithms like dynamic programming. According to recent advances in algorithmic alignment, neural networks with structured computation procedures can be trained efficiently. We demonstrate the potential of graph neural network in supporting sample efficient learning by showing that Deep Graph Value Network can outperform unstructured baselines by a large margin in solving Markov Decision Process (MDP). We believe this would open up a new avenue for structured agents design. See https://github.com/drmeerkat/Deep-Graph-Value-Network for the code.
\end{abstract}

\section{Introduction}

Deep reinforcement-learning algorithms have produced breakthroughs in recent years.
However, agents with powerful non-linear function approximators also require large amounts of experience to learn. In this work, the question we focus on is how to get the best of both neural networks and non-neural network algorithms (e.g. Dynamic Programming) to learn effectively.

One intuitive solution is to execute those algorithms with neural networks. Different datasets favor different network structures and computational process. Recent work has shown that better algorithmic alignment improves sample complexity and generalization~\citep{gnndp}. Given the fact that GNNs align well with dynamic programming (DP), we presume value iteration, as a probabilistic version of Bellman-ford algorithm, should also be solvable with GNNs. Thus, we propose Deep Graph Value Network (DeepGV), a message-passing framework, to robustly solve the given MDP. We empirically verify its effectiveness as a general MDP solver and its potential towards building general structured agents.

\section{Algorithmic Alignment}

We adopt the theoretical framework of \citep{gnndp}. In this section, we will briefly review the basic definitions. For more details, please refer to \citet{gnndp}.

\noindent \textbf{Definition 2.1. (PAC learning and sample complexity).} Fix an error parameter $\epsilon > 0$ and failure probability $\delta \in (0,1)$. Suppose $\{x_i, y_i\}^M_{i=1}$ are i.i.d. samples from distribution $\mathcal{D}$, and the data satisfies $y_i = g(x_i)$ for some underlying function $g$. Let $f = \mathcal{A}\left( \{x_i, y_i\}_{i=1}^M \right)$ be the function generated by a learning algorithm $\mathcal{A}$. Then, $g$ is \emph{$(M, \epsilon, \delta)$-learnable} with $\mathcal{A}$ if
\begin{align}
    \mathbb{P}_{x\in\mathcal{D}}[||f(x)-g(x)|| \leq\epsilon] \geq 1-\delta .
\end{align}
The \emph{sample complexity}  $\mathcal{C}_{\mathcal{A}}(g,\epsilon,\delta)$ is
\begin{align}
     \min \{M|\mathbb{P}_{x\in\mathcal{D}}[||f(x)-g(x)|| \leq\epsilon] \geq 1-\delta\} .
\end{align}

\noindent \textbf{Definition 2.2. (Algorithmic alignment).} Let $g$ be a reasoning function and $\mathcal{F}$ a neural network with $n$ modules $f_i$. The module functions $g_1, ..., g_n$ generate $g$ for $\mathcal{F}$ if, by replacing $f_i$ with $g_i$, the network simulates $g$. Then, $\mathcal{F}(M, \epsilon,\delta)$-algorithmically aligns with $g$ if (1) $g_1, ..., g_n$ generate $g$ and (2) there are learning algorithms $\mathcal{A}_i$ for the $f_i$ such that $n\cdot \max_{i}\mathcal{C}_{\mathcal{A}_i}(g_i, \epsilon, \delta) \leq M$.

From the definition, it is clear that only if each module of the neural network aligns well with the underlying function modules can we reduce $M$ to its minimum.

\section{Deep Graph Value Network}

\begin{figure}%
    \centering
    \includegraphics[width=0.4\textwidth]{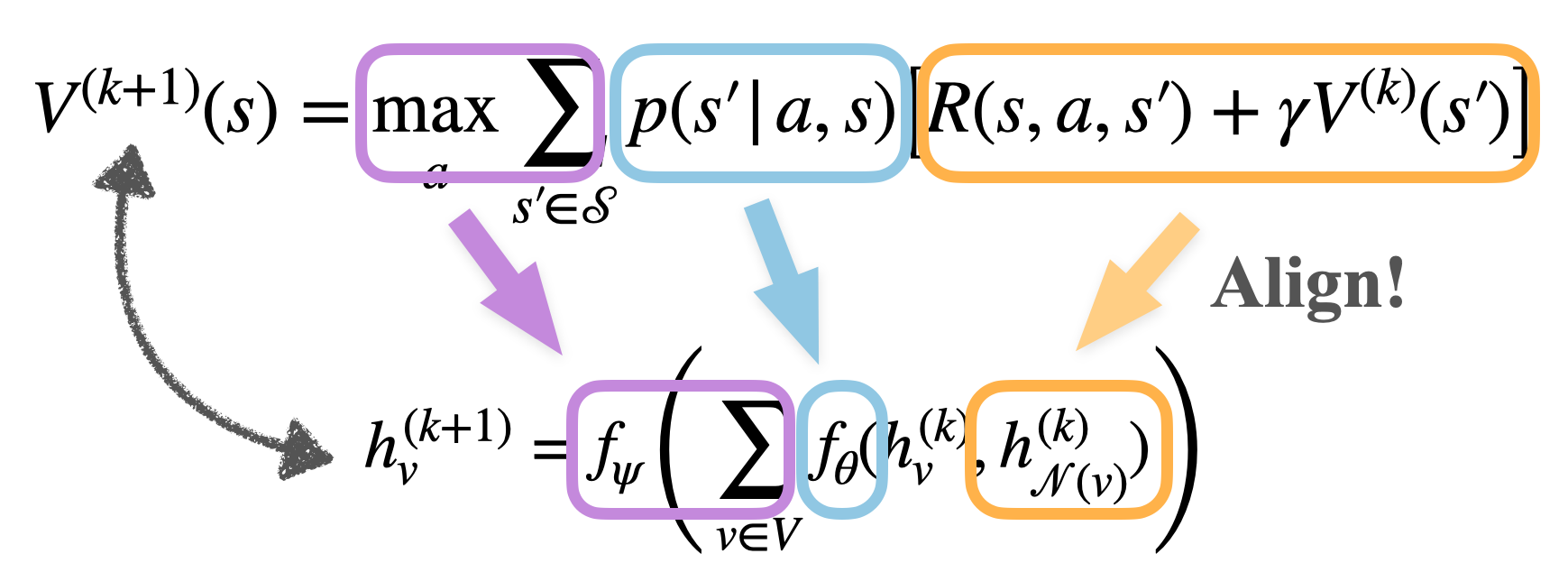}
    \caption{An example of how generally GNNs align with value iteration. Both GNNs and VI share the same loop structure, which does not need to be learned. Thus, it would be relatively simpler for GNNs than generic Multi-layer Perceptrons (MLPs) to learn to execute value iteration.}%
    \label{fig:align}%
\end{figure}

Due to the space limitation, for general context of MDP and graph neural network, please refer to \citet{sutton2018reinforcement, inductive_bias}. 
In each iteration of message distribution and aggregation, every node first updates its embeddings according to information from its neighbours and edges between them. After some iteration of execution, we aggregate all the information and extract answers from it with another network. As shown in Fig.~\ref{fig:align}, comparing general GNN updates with value iteration, they share quite similar underlying computational steps, which can be simulated by neural networks easily. 

The framework of Deep Graph Value Network is as follows. The network takes a complete graph $G(V, E)$ as input with different attributes assigned to nodes and edges. Each node represents a state in the given MDP $\langle S, A, R, \gamma, T\rangle$. Edges between nodes represent actions with rewards binding to them.   

In iteration $k+1$, the node embedding $h_i^{(k+1)}$ is calculated as follows, 
\begin{align}
    f_{\psi}\left ( \cat_{a\in\mathcal{A}} \sum_{j \in \mathcal{N}(i)} A_{i,j}^{(k+1)}\left(r(i, a,j)+\gamma h_j^{(k)}\right) \right ) ,
\end{align}
where $\cat(\cdot)$ is a concatenate operator, $r$ is the embedded reward attributes attached to edges, $\gamma$ is the discount factor and $A_{i,j}$ is the transition probability learned by a self-attention mechanism. The part inside the square brackets calculates the Q-value for state $i$ and the local aggregate function $f_{\gamma}$ learns to execute $\max$. The attention mechanism is learned as follows,
\begin{align}
    A_{i, j}^{(k+1)} &= \frac{exp(e_{i,j}^{(k+1)})}{\sum_{l\in \mathcal{N}(i)}exp(e_{i,l}^{(k+1)})} ,\\
    a_{i,j}^{(k+1)} &= f_{\theta}(h_i^{(k)},  h_i^{(k)}) .
\end{align}

After $K$ iterations, we aggregate information from all the nodes and pass them through an MLP to get value-ranking predictions:
\begin{align}
    \hat{y} = f_{\omega}\left(\cat_{i\in V}h_i^{(K)}\right) .
\end{align}
 



\begin{figure}%
    \centering
    \begin{subfigure}[t]{\linewidth}
        \centering
        \includegraphics[width=\textwidth]{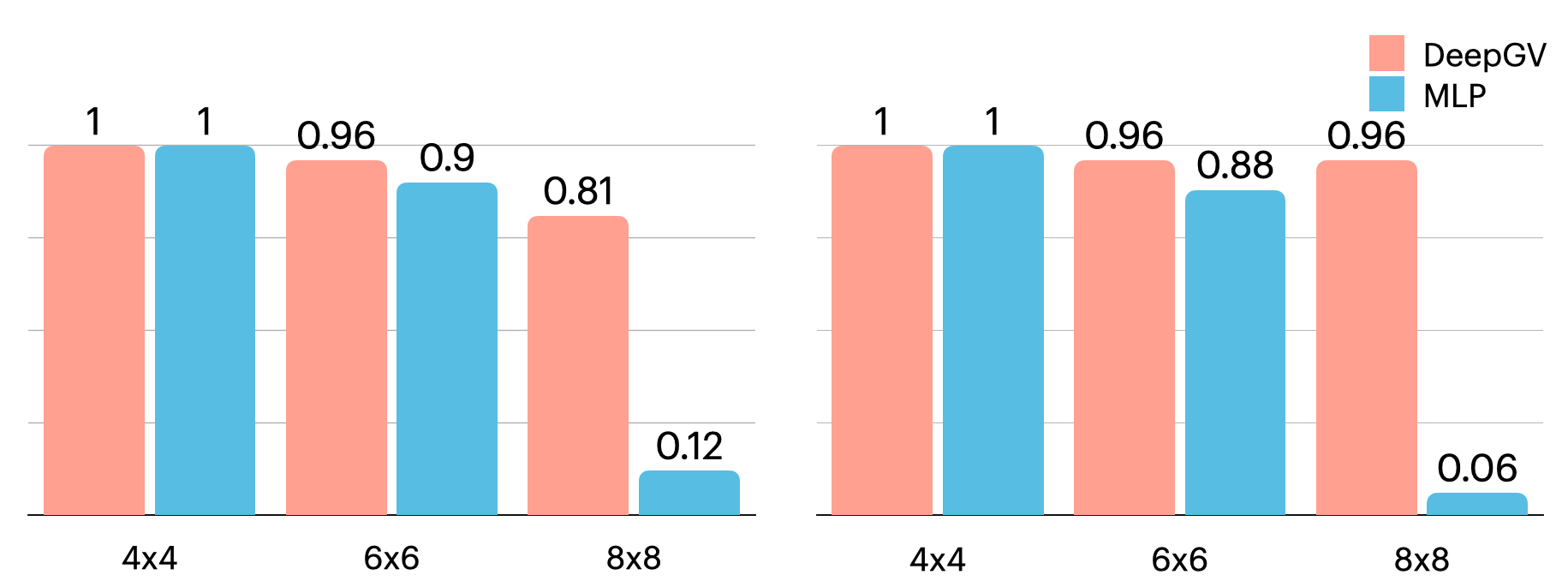}
        \caption{Value ranking accuracy varying grid world size.}
    \end{subfigure}
    \newline
    %
    \begin{subfigure}[t]{\linewidth}
        \centering
        \includegraphics[width=\textwidth]{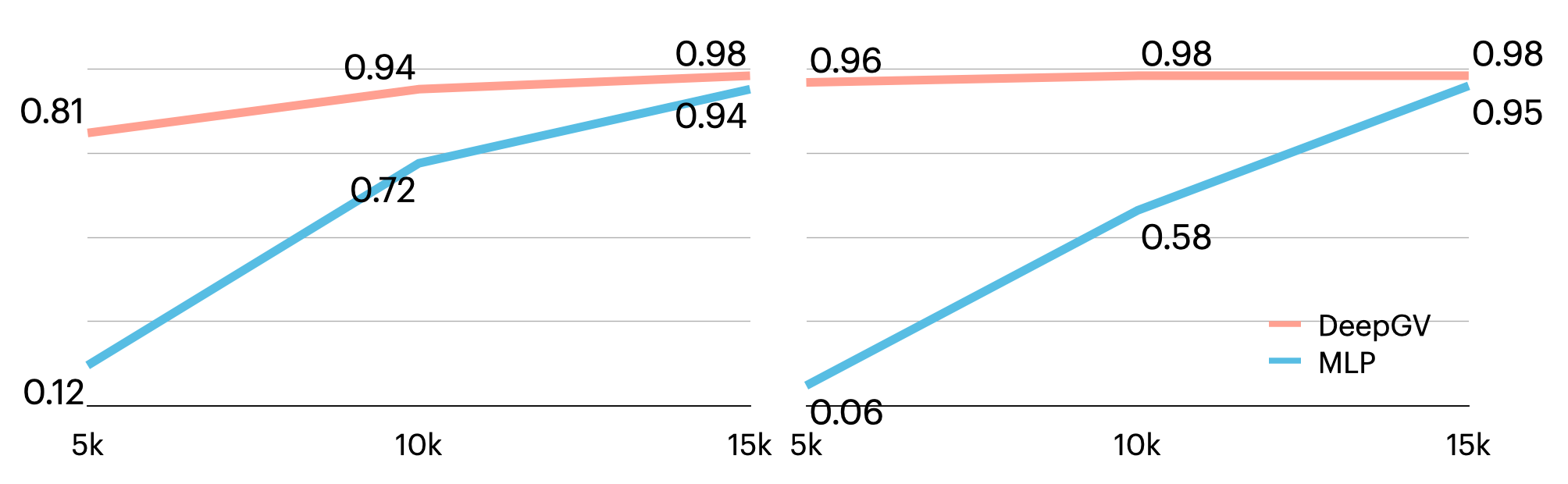}
        \caption{Training set size vs. test accuracy.}
    \end{subfigure}%
    \caption{Test accuracy and sample complexity on two types of grid world. In each sub-figure, the LHS result comes from plain grid world while the RHS result comes from grid world with traps. (a) MLP fails on $8\times8$ world while DeepGV performs well. (b) As the training set size increases, DeepGV achieves good performance with fewer training examples.}
    \label{fig:result}%
\end{figure}

\section{Experiments and Discussion}

In this section, we apply our framework to solve MDPs and compare its performance with that of vanilla MLPs. Intuitively, unlike DeepGV, MLP doesn't align well with dynamic programming. Thus, we expect to see that DeepGV can easily be trained and outperforms the MLP by a large margin. 


We evaluate the networks on two different types of MDPs. The first one is a slippery $N \times N$ grid world without obstacles and fire states. The second type has obstacles and fire states. We generate different instances of these two kinds of MDPs and evaluate two algorithms on each. 
To solve a given MDP, the input is randomly initialized state values, state coordinates and the queried state coordinates. The expected output is the value rank of the queried state. 


As shown in fig.~\ref{fig:result}, when we fix the training set size and increase the task difficulty (the size of grid world), DeepGV can still be trained to a satisfying level while MLP performance drops severely in the $8\times8$ grid world. When we varied the training set size and fixed the size of the grid world at $8\times8$, there is a clear gap between MLP and DeepGV performance. These results empirically support Theorem 3.6 from \cite{gnndp} that better algorithmic alignment induces better sample complexity. Overall, our experiment verifies the intuition that DeepGV can be trained easily and outperforms the MLP by a large margin.


In conclusion, we presented Deep Graph Value Network (DeepGV), a graph network structure that can efficiently learn to solve MDPs by executing a similar computational procedure to that of value iteration. 
Our preliminary experimental results show the performance gap between structured networks and unstructured networks in solving relational tasks suggesting the promising potential of message-passing mechanisms in building structured RL agents. We believe this opens an exciting new avenue for future work on designing high-efficiency agents incorporating algorithmic prior.

{ 
\bibliography{reference}}

\begin{thebibliography}{3}
\providecommand{\natexlab}[1]{#1}
\providecommand{\url}[1]{\texttt{#1}}
\providecommand{\urlprefix}{URL }
\expandafter\ifx\csname urlstyle\endcsname\relax
  \providecommand{\doi}[1]{doi:\discretionary{}{}{}#1}\else
  \providecommand{\doi}{doi:\discretionary{}{}{}\begingroup
  \urlstyle{rm}\Url}\fi

\bibitem[{Battaglia, Hamrick et~al.(2018)}]{inductive_bias}
Battaglia, P.~W.; Hamrick, J.~B.; et~al. 2018.
\newblock Relational inductive biases, deep learning, and graph networks.
\newblock \emph{CoRR} abs/1806.01261.

\bibitem[{Sutton and Barto(2018)}]{sutton2018reinforcement}
Sutton, R.~S.; and Barto, A.~G. 2018.
\newblock \emph{Reinforcement learning: An introduction}.
\newblock MIT press.

\bibitem[{Xu, Li et~al.(2020)}]{gnndp}
Xu, K.; Li, J.; et~al. 2020.
\newblock What Can Neural Networks Reason About?
\newblock In \emph{ICLR}.

\end{thebibliography}

\end{document}